\documentclass[11pt,a4paper]{article}
\usepackage[hyperref]{emnlp2020}
\usepackage{times}
\usepackage{latexsym}
\usepackage{tabu}
\usepackage{amsmath}
\usepackage{amsfonts}
\usepackage{dsfont}
\usepackage{pifont}
\usepackage{booktabs}
\usepackage{multirow}
\usepackage{xcolor,colortbl}
\usepackage{graphicx}

\usepackage{microtype}

\aclfinalcopy 

\definecolor{Gray}{gray}{0.9}

\newcommand\tn[1]{\textnormal{#1}}
\newcommand\bs[1]{\boldsymbol{#1}}

\title{Multi-Referenced Training for Dialogue Response Generation}

\author{Tianyu Zhao \\
Graduate School of Informatics \\
Kyoto University \\ 
{\tt zhaoty.ting@gmail.com}
\And
Tatsuya Kawahara \\
Graduate School of Informatics \\
Kyoto University \\
{\tt kawahara@i.kyoto-u.ac.jp}
}

\date{}

\begin{document}
\maketitle

\begin{abstract}
  In open-domain dialogue response generation, a dialogue context can be continued with diverse responses, and the dialogue models should capture such one-to-many relations. In this work, we first analyze the training objective of dialogue models from the view of Kullback\textendash Leibler divergence (KLD) and show that the gap between the real world probability distribution and the single-referenced data's probability distribution prevents the model from learning the one-to-many relations efficiently. Then we explore approaches to multi-referenced training in two aspects. Data-wise, we generate diverse pseudo references from a powerful pretrained model to build multi-referenced data that provides a better approximation of the real-world distribution. Model-wise, we propose to equip variational models with an expressive prior, named linear Gaussian model (LGM). Experimental results of automated evaluation and human evaluation show that the methods yield significant improvements over baselines. We will release our code and data in \url{https://github.com/ZHAOTING/dialog-processing}.
\end{abstract}

\section{Introduction}
\label{sec:intro}

\begin{figure}[!t]
  \centering
  \includegraphics[width=1.0\columnwidth]{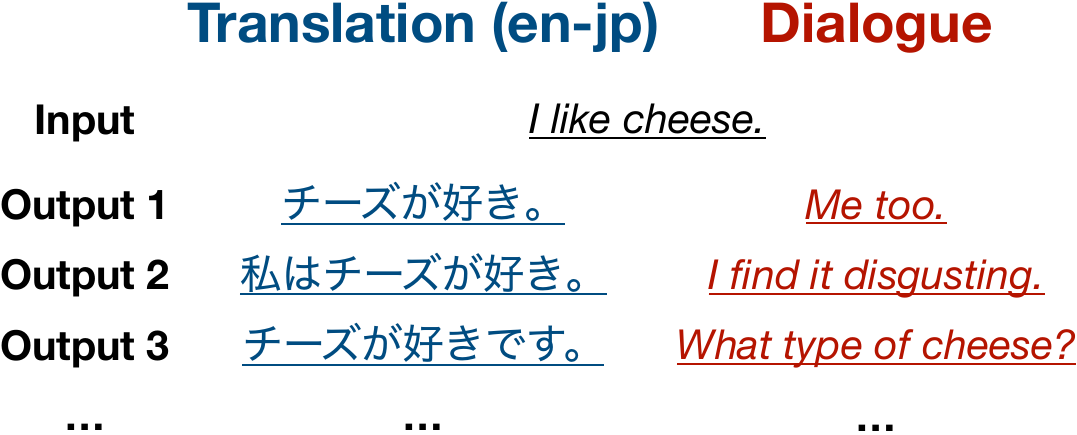}
  \caption{Examples of multiple valid outputs given the same input in machine translation and dialogue.}
  \label{fig:mt_vs_dial}
\end{figure}

Open-domain dialogue modeling has been formulated as a seq2seq problem since \citet{ritter2011data} and \citet{vinyals2015neural} borrowed machine translation (MT) techniques~\citep{koehn2007moses, sutskever2014sequence} to build dialogue systems, where a model learns to map from \textit{one} context to \textit{one} response. In MT, \textit{one}-to-\textit{one} mapping is a reasonable assumption since an MT output is highly constrained by its input. Though we may use a variety of expressions to translate the same input sentence, these different translations still highly overlap with each other lexically and semantically (see the translation example in Figure~\ref{fig:mt_vs_dial}), and learning from one output reference is often sufficient for training a good MT system~\citep{kim2016sequence}. In dialogues, however, the same input can be continued with multiple diverse outputs which are different in both the used lexicons and the expressed semantic meanings (see the dialogue example in Figure~\ref{fig:mt_vs_dial}). Learning from barely one output reference ignores the possibility of responsding with other valid outputs and is thus insufficient for building a good dialogue system. 

The current dialogue modeling paradigm is largely derived from MT research, and it trains dialogue models with one output reference given each input. In this paper, we will investigate why single-referenced training harms our dialogue models and how to apply multi-referenced training.

\section{Why Multi-Referenced Training Matters?}
\label{sec:kld}

A dialogue context $X$ can be continued with a set of different responses $\{ Y_1, \cdots, Y_i, \cdots \}$. In the training of a response generation model, we expect to model the \textit{real probability distribution} $P(\bs{Y}|X)$ with \textit{model probability distribution} $P_{\theta}(\bs{Y}|X)$ for each context $X$, where $\theta$ is the model parameters. In most scenarios, however, we can only rely on a data set $D = \{ (X^{(j)}, Y^{(j)}_{1}) \}_{j}^{|D|}$,\footnote{For simplicity, we define a response in $D$ as the first response to its context, and thus its subscript is 1. We will omit the superscript in the rest of the paper.} where only one valid response is presented. This results in a \textit{data probability distribution} $P_{\tn{D}}(\bs{Y}|X)$ that is very different from $P(\bs{Y}|X)$. In fact, $P_{\tn{D}}(\bs{Y}|X)$ is an one-hot vector where the first element is 1 while others are 0.

\paragraph{Emprical training objective} As a result, we optimize a model to match the \textit{model probability distribution} and the \textit{data probability distribution}. From the view of Kullback\textendash Leibler divergence (KLD), we can see it as to minimize $D_{\tn{KL}}( P_{\tn{D}} || P_{\theta})$:
\begin{align*}
  - \begin{matrix} \sum_{i} \end{matrix} P_{\tn{D}}(Y_i | X) \log \tfrac{P_{\theta}(Y_i | X)}{P_{\tn{D}}(Y_i | X)},
\end{align*}
which is identical to minimize the following target function after ignoring terms that are not related to the model parameter $\theta$:
\begin{align*}
  \mathcal{L}_{\tn{D}}(X, \bs{Y}) & = - \begin{matrix} \sum_{i} \end{matrix} P_{\tn{D}}(Y_i | X) \log P_{\theta}(Y_i | X) \\
  & = - \begin{matrix} \sum_{i} \end{matrix} \mathds{1}\{ i=1 \} \log P_{\theta}(Y_i | X) \\
  & = - \log P_{\theta}(Y_1 | X). \label{equ:sent_nll}
\end{align*}
The resulting objective is the negative log likelihood (NLL) loss function commonly used in the implementation of dialogue models.

\paragraph{Ideal training objective} We hope to minimize the KLD between the \textit{model probability distribution} and the \textit{real probability distribution}, $D_{\tn{KL}}( P || P_{\theta})$:
\begin{align*}
  - \begin{matrix} \sum_{i} P(Y_i | X) \log \tfrac{P_{\theta}(Y_i | X)}{P(Y_i | X)} \end{matrix},
\end{align*}
which is identical to minimize:
\begin{align*}
  \mathcal{L}^{*}(X, \bs{Y}) = - \begin{matrix} \sum_{i} P(Y_i | X) \log P_{\theta}(Y_i | X) \end{matrix}.
\end{align*}
However, $\mathcal{L}^{*}$ is intractable because 1)~there are often an enormous number of valid responses, and 2)~we cannot obtain the real probability of a certain response $P(Y_i | X)$.

\paragraph{The problem and proposed solutions} The gap between $\mathcal{L}_{\tn{D}}$ and $\mathcal{L}^{*}$ is caused by the difference between $P_{\tn{D}}(\bs{Y} | X)$ and $P(\bs{Y} | X)$, and it prevents dialogue models from learning one-to-many mappings efficiently. To alleviate this problem, we propose methods to allow for multi-referenced training in two aspects.
\begin{itemize}
  \item Data-wise, we replace the original data distribution $P_{\tn{D}}(\bs{Y} | X)$ with an approximated real distribution $P_{\phi}(\bs{Y} | X)$ by generating up to 100 pseudo references from a \textit{teacher model} parameterized by $\phi$. We show that using the newly created data yields significant improvement.
  \item Model-wise, we argue that a model requires an encoder of large capacity to capture sentence-level diversity, and thus we propose to equip the variational hierarchical recurrent encoder-decoder (VHRED) model with a linear Gaussian model (LGM) prior. The proposed model outperforms VHRED baselines with unimodal Gaussian prior and Gaussian Mixture Model (GMM) prior in evaluation experiments.
\end{itemize}

\section{Related Works}
\label{sec:related}

\subsection{Knowledge Distillation}
In the context of machine translation, \citet{kim2016sequence} proposed that a \textit{teacher model}'s knowledge can be transferred to a \textit{student model} on a sequence level. They showed that transferring sequence-level knowledge is roughly equal to training on sequences generated by the \textit{teacher model} as references. However, \textit{one} generated reference given each input is sufficient for transferring the teacher's MT knowledge, while we will show in following experiments that training with multiple generated references can yield far better results in dialogue response generation. This confirms our earlier hypothesis that the one-to-many nature is an important characteristic that distinguishes open-domain dialogue modeling from other tasks such as machine translation.

In task-oriented dialogues, \citet{peng2019teacher} proposed to transfer knowledge from multiple teachers for multi-domain task-oriented dialogue response generation via policy distillation and word-level output distillation. \citet{tan2018multilingual} applied a similar approach to multilingual machine translation. \citet{kuncoro2019scalable} transferred syntactic knowledge from recurrent neural network grammar (RNNG, \citealp{dyer2016recurrent}) models to a sequential language model.

\subsection{Data Augmentation and Manipulation}
The multi-referenced training approach can be seen as a data augmentation method. Prior works on data augmentation in text generation tasks often operate on word level while our method performs sentence-level augmentation. \citet{niu2019automatically} proposed to apply semantic-preserving perturbations to input words for augmenting data in dialogue tasks. \citet{zheng2018multi} investigated generating pseudo references by compressing existing multiple references into a lattice and picking new sequences from it. \citet{hu2019learning} used finetuned BERT~\citep{devlin2019bert} as the data manipulation model to generate word substitutions via reinforcement learning.

Another line of research focuses on filtering high-quality training examples for dialogue response generation. \citet{csaky2019improving} proposed to remove generic responses using an entropy-based approach. \citet{shang2018learning} trained a data calibration network to assign higher instance weight to more appropriate responses.

\subsection{Expressive Dialogue Models}
\label{ssec:expressive_models}
Besides manipulating the training data, dialogue researchers have attempted to strengthen dialogue models' capacity for capturing complex relations between the input context and the output responses. \citet{zhou2017mechanism} incorporated mechanism embeddings $\mathbf{m}$ into a seq2seq model for dialogue response generation. The mechanism-aware model decodes a response by selecting a mechanism embedding $\mathbf{m_k}$ and combining it with context encoding $\mathbf{c}$. Therefore, the model is capable of generating diverse responses by choosing different mechanisms. \citet{zhang2018tailored} borrowed the conditional value-at-risk (CVaR) from finance as an alternative to sentence likelihood (which is negated $\mathcal{L}_{\tn{D}}$) for optimization. Optimizing the CVaR objective can be seen as rejecting to optimize on easy instances whose model probabilities are larger than a threshold $\alpha$. \citet{qiu2019training} proposed a two-step VHRED variant for modeling one-to-many relation. In the first step, they forced the dialogue encoding vector $\mathbf{c}$ to store common features of all response hypotheses $Y_{2:N+1}$ by adversarial training. In the second step, they trained the latent variable $\mathbf{z}$ to capture response-specific information by training with a multiple bag-of-words (MBoW) loss. These three methods will be compared with the proposed model in this work as they have focused on modeling one-to-many relations in dialogue response generation.

\citet{gao2019generating} relied on vocabulary prediction to model sentence-level discrepancy. \citet{chen2019generating} utilized a mechanism-based architecture and proposed a posterior mapping method to select the most proper mechanism. \citet{gu2019dialogwae} proposed to train latent dialogue models in the framework of generative adversarial network (GAN). They optimized the model by minimizing the distance between its prior distribution and its posterior distribution via adversarial training.

\section{Preliminary}
\label{sec:pre}

\subsection{Models}
\paragraph{HRED} We use the hierarchical recurrent encoder decoder (HRED, \citealp{serban2016building}) as the baseline model, where a hierarchical RNN-based encoder $\mathcal{E_{\theta}}(\cdot)$ encodes the context $X$ and produces an encoding vector $\mathbf{c}$, and an RNN-based decoder $\mathcal{D_{\theta}}(\cdot)$ takes $\mathbf{c}$ as input and computes the conditional probability of a response $P_{\theta}(Y_i | X)$ as the product of word probabilities.
\begin{align*}
  \mathbf{c} &= \mathcal{E_{\theta}}(X) \\
  P_{\theta}(Y_i | X) &= \begin{matrix} \prod_{l=1}^L P_{\theta}(Y_{i,l} | Y_{i,:l-1}, X) \end{matrix} \\
    &= \begin{matrix} \prod_{l=1}^L \mathcal{D_{\theta}}(Y_{i,l} | Y_{i,:l-1}, \mathbf{c}) \end{matrix},
\end{align*}
where $Y_{i,j}$ stands for the $j$-th word in $Y_i$ and $L$ is the length of $Y_i$.

\paragraph{VHRED}
For a given context, the HRED produces a fixed-length encoding vector $\mathbf{c}$ and relies on it to decode various responses. However, the one-to-many mapping in dialogues is often too complex to capture with a single vector $\mathbf{c}$. \citet{serban2017hierarchical} proposed variational HRED (VHRED) and used a stochastic latent variable $\mathbf{z}$ that follows a multivariate Gaussian distribution to strengthen the model's expressiveness.
\begin{align*}
  \boldsymbol{\mu}, \boldsymbol{\sigma} &= \tn{MLP}_{\theta}(\mathbf{c}) \\
  \mathbf{z} &\sim \tn{Gaussian}(\boldsymbol{\mu}, \boldsymbol{\sigma}^2\boldsymbol{I}) \\
  P_{\theta}(Y_i | X) &= \begin{matrix} \prod_{l=1}^L \mathcal{D_{\theta}}(Y_{i,l} | Y_{i,:l-1}, \mathbf{c}, \mathbf{z}) \end{matrix},
\end{align*}
where $\boldsymbol{\mu}$ and $\boldsymbol{\sigma}^2\boldsymbol{I}$ are parameters of the Gaussian distribution. In order to mitigate the infamous \textit{posterior collapse} problem in variational models, it is common to apply tricks such as annealing KLD loss~\citep{bowman2016generating} and minimizing a bag-of-words (BoW) loss~\citep{zhao2017learning}.

\paragraph{VHRED with GMM prior}
\citet{gu2019dialogwae} showed that the performance of the vanilla VHRED is limited by the single-modal nature of Gaussian distribution, and thus they proposed to use as prior a Gaussian Mixture Model (GMM) with $K$ components to capture multiple modes in $\mathbf{z}$'s probability distribution, such that $\mathbf{z}$ is sampled in the following way:
\begin{align*}
  \boldsymbol{\mu}_k, \boldsymbol{\sigma}_k, \pi_k &= \tn{MLP}_{\theta, k}(\mathbf{c}) \\
  \mathbf{z} &\sim \tn{GMM}(\{\boldsymbol{\mu}_k, \boldsymbol{\sigma}_k^2\boldsymbol{I}, \pi_k\}_{k=1}^{K}),
\end{align*}
where $\pi_k$ is the weight of the $k$-th component. We refer to the VHRED with $K$-component GMM prior as VHRED$_{\textit{gmm}K}$.

\paragraph{GPT2} We finetune a pre-trained medium-sized GPT2~\citep{radford2019language} on dialogues and use it as the \textit{teacher model} to obtain $P_{\phi}(\bs{Y} | X)$ as an approximation of $P(\bs{Y} | X)$. GPT2 has been shown to reach low perplexity on real-world texts, and it can generate high-quality responses~\citep{wolf2019transfertransfo, zhang2019dialogpt}. Therefore, we expect it to provide a relatively accurate approximation of the real-world distribution.

\subsection{Data}
\label{ssec:data}
We use the DailyDialog corpus~\cite{li2017dailydialog} to investigate the effects of the proposed methods. We make a roughly 0.8:0.1:0.1 session-level split for training, validation, and test, respectively.\footnote{See the Appendix for more details about the data set.}

\subsection{Metrics}
\label{ssec:metrics}
\paragraph{Automated Metrics} We use perplexity on the test data as the metric for intrinsic evaluation. For extrinsic evaluation, we choose BLEU-2 and three types of word embedding similarities (Embedding Extrema, Embedding Average, Embedding Greedy) to measure the closeness between a hypothesis and the corresponding ground-truth reference.

\paragraph{Dialogue Response Evaluator} Besides the automated metrics above, we also use RoBERTa-eval, a model-based dialogue response evaluator, to approximate human judgement~\citep{zhao2020designing}. RoBERTa-eval computes the appropriateness (a real value from 1 to 5) of a response hypothesis by conditioning on its context instead of by comparing with its reference. It has been shown to correlate with human judgement significantly better than automated metrics. The authors reported Pearson's $\rho$ = 0.64 and Spearman's $\rho$ = 0.66 on the DailyDialog corpus.

\paragraph{Human Evaluation} Following~\citet{adiwardana2020towards}, we ask Amazon MTurk human annotators to evaluate each response on two criteria, sensibleness and specificity. Both metrics take binary values, and we use their average (knowns as Sensibleness and Specificity Average, SSA) to assess the overall quality.

\section{Proposal: Enhancing Data for Multi-Referenced Training}
\label{sec:data_multi}

To enhance the training data, we try to close the gap between $P_{\tn{D}}(\bs{Y} | X)$ and $P(\bs{Y} | X)$. Since all probability mass is on a single response in $P_{\tn{D}}(\bs{Y} | X)$, the gap can be closed by assigning some mass to other valid responses. We use a finetuned GPT2$_{\textit{md}}$ to generate $N$ hypotheses as valid responses, and let the probability mass to be assigned to them uniformly. It results in $P_{\phi}(\bs{Y} | X)$ wherein $N$ elements have $\tfrac{1}{N}$ probability. The new training objective is:
\begin{align*}
  \tilde{\mathcal{L}^{*}}(X, \bs{Y}) & = - \tfrac{1}{N} \begin{matrix} \sum_{i=2}^{N+1} \log P_{\theta}(Y_i | X) \end{matrix},
\end{align*}
where we assume responses $Y_2$ to $Y_{N+1}$ are generated responses. 

Training with the new loss function can be achieved by directly replacing the ground-truth responses in the training data with the hypotheses.\footnote{We will refer to the original response as ground truth and the generated responses as hypotheses. A reference can be either a ground-truth response or a hypothesis response.}

Sequences generated by beam search often highly overlap both lexically and semantically~\citep{li2016diversity}. Therefore, we use nucleus sampling with top probability 0.95~\citep{holtzman2019curious} to generate 100 hypotheses as for each context in the training data. 

\begin{table*}[!t]
  \centering
  \begin{tabu} to 1.0\textwidth {lrcccccc}
      \toprule
      \multirow{2}{*}{\textbf{Model}} & \multirow{2}{*}{\textbf{Data}} & \multirow{2}{*}{\textbf{ppl.}} & \multirow{2}{*}{\textbf{BLEU-2}} & \multicolumn{3}{c}{\textbf{Embedding Similarity}} & \multirow{2}{*}{\textbf{Reval}} \\
      & & & & \textbf{Ext} & \textbf{Avg} & \textbf{Grd} & \\
      \midrule
      \multicolumn{8}{c}{\textit{Single-referenced training (baseline w/o KD)}} \\
      \midrule
      HRED & 1 GT & 29.00 & 6.46 & 39.40 & 60.80 & 43.92 & 3.42 \\
      \midrule
      \multicolumn{8}{c}{\textit{Single-referenced training (baseline token-KD, \textnormal{\S\ref{ssec:token_kd}})}} \\
      \midrule
      HRED$_{\textit{token-KD}}$ & 1 GT & 27.68 & 6.90 & 39.83 & 62.33 & 45.11 & 3.45 \\
      \midrule
      \multicolumn{8}{c}{\textit{Single-referenced training (baseline seq-KD, \textnormal{\S\ref{ssec:hyp_data}})}} \\
      \midrule
      HRED & 1 hyp & 35.08 & 6.62 & 39.66 & 61.96 & 44.75 & 3.61 \\
      \midrule
      \multicolumn{8}{c}{\textit{Multi-referenced training (proposed seq-KD, \textnormal{\S\ref{ssec:hyp_data}})}} \\
      \midrule
      \multirow{3}{*}{HRED} & 5 hyp & 23.10 & 7.13 & 40.23 & 62.43 & 45.44 & 3.82 \\
      & 20 hyp & 21.15 & \textbf{7.38} & \textbf{40.52} & \textbf{62.53} & \textbf{45.64} & 3.87 \\
      & 100 hyp & \textbf{20.93} & 7.28 & 40.26 & 62.22 & 45.30 & \textbf{3.89} \\
      \bottomrule
  \end{tabu}
  \caption{Experimental results of data enchancement for multi-referenced training. \textbf{GT}~\textendash~ground truth; \textbf{hyp}~\textendash~hypotheses; \textbf{ppl}~\textendash~perplexity; \textbf{Ext}~\textendash~Embedding Extrema; \textbf{Avg}~\textendash~Embedding Average; \textbf{Grd}~\textendash~Embedding Greedy; \textbf{Dist}-$n$~\textendash~Distinct-$n$; \textbf{Reval}~\textendash~RoBERTa-eval score. Grey background is for distinguishing areas of different models.}
  \label{tab:data_result}
\end{table*}

\subsection{Training with Hypotheses}
\label{ssec:hyp_data}

In this part, we compare baseline HRED models trained with only ground truth (GT) and with different numbers of hypotheses. Since using $N$ hypotheses makes the training data $N$ times larger, we accordingly adjust the maximum number of training epochs. We found that all the models can converge in the given epochs.~\footnote{See the Appendix for experimental settings and statistics of model size and training cost.}

As shown in Table~\ref{tab:data_result}, replacing 1 GT with 1 hypothesis yields a boost on most metrics. Further increasing the number of hypotheses will continue to improve the model's performance. It is worth noting that when the number of hypotheses is increased from 20 to 100, the performance gain is limited. This suggests that as training data increases, the model's capacity might have become a bottleneck.

\subsection{Comparing with Knowledge Distillation}
\label{ssec:token_kd}

The proposed data enhancement can be considered as a multi-sequence sequence-level knowledge distillation (seq-KD), and it has been shown to significantly outperform single-sequence seq-KD (i.e. the 1 hyp setting). We would also like to compare it with token-level KD (token-KD), where the student HRED learns to match its softmax output with the teacher GPT2 on every token in the output sequence~\citep{kim2016sequence}. The model is referred to as HRED$_{\textit{token-KD}}$.

While token-KD outperforms single-sequence seq-KD in some metrics according to Table~\ref{tab:data_result}, the proposed multi-sequence seq-KD is much better than token-KD in all metrics. Other drawbacks of token-KD include: 1) It requires the student model to have the same vocabulary as the teacher model; 2) The teacher model has to make inference to provide a probability distribution over the vocabulary for every output token and thus makes the training extremely slow.

\section{Proposal: Enhancing Model for Multi-Referenced Training}
\label{sec:model_multi}

We have previously seen the HRED's performance gain when we increase the number of hypotheses from 1 to 20, but it starts to degrade when we further increase the number to 100. A conjecture is that the model's capacity is insufficient to learn too complex input-output relations.

\begin{table*}[!t]
  \centering
  \begin{tabu} to 1.0\textwidth {lrcccccc}
      \toprule
      \multirow{2}{*}{\textbf{Model}} & \multirow{2}{*}{\textbf{Data}} & \multirow{2}{*}{\textbf{ppl.}} & \multirow{2}{*}{\textbf{BLEU-2}} & \multicolumn{3}{c}{\textbf{Embedding Similarity}} & \multirow{2}{*}{\textbf{Reval}} \\
      & & & & \textbf{Ext} & \textbf{Avg} & \textbf{Grd} & \\
      \midrule
      \multicolumn{8}{c}{\textit{Baseline model}} \\
      \midrule
      HRED & 100 hyp & 20.93 & 7.28 & 40.26 & 62.22 & 45.30 & 3.89 \\
      \midrule
      \multicolumn{8}{c}{\textit{Baseline larger model} (\S\ref{ssec:larger_model}) } \\
      \midrule
      HRED$_{\textit{l}}$ & 100 hyp & 20.81 & 7.36 & 40.66 & 62.53 & 45.48 & 3.90 \\
      HRED$_{\textit{xl}}$ & 100 hyp & \textbf{20.69} & 7.21 & 40.43 & 62.51 & 45.65 & 3.85 \\
      \midrule
      \multicolumn{8}{c}{\textit{Baseline variational model} (\S\ref{ssec:variational_model}) } \\
      \midrule
      VHRED & 100 hyp & 56.54 & 5.39 & 38.49 & 62.38 & 44.59 & 3.25 \\
      VHRED$_{\textit{gmm}5}$ & 100 hyp & 50.44 & 5.44 & 38.77 & 62.55 & 44.79 & 3.33 \\
      \midrule
      \multicolumn{8}{c}{\textit{Proposed variational model} (\S\ref{ssec:lgm}) } \\
      \midrule
      \multirow{5}{*}{VHRED$_{\textit{lgm}5}$} & 1 GT & 39.97 & 6.10 & 40.30 & 64.03 & 45.92 & 3.33 \\
      & 1 hyp & 50.44 & 6.12 & 40.26 & 64.17 & 46.05 & 3.50 \\
      & 5 hyp & 30.85 & 6.61 & 41.31 & 65.31 & 47.19 & 3.73 \\
      & 20 hyp & 29.74 & 6.82 & 41.33 & 65.29 & 47.39 & 3.76 \\
      & 100 hyp & 28.76 & 6.79 & 41.31 & 65.18 & 47.19 & 3.76 \\
      \\
      \multirow{5}{*}{VHRED$_{\textit{lgm}20}$} & 1 GT & 46.46 & 6.70 & 41.12 & 64.98 & 46.83 & 3.64 \\
      & 1 hyp & 46.45 & 6.65 & 41.10 & 64.95 & 46.77 & 3.64 \\
      & 5 hyp & 29.18 & 6.99 & 41.80 & 65.72 & 47.68 & 3.82 \\
      & 20 hyp & 26.93 & 7.07 & 42.29 & 66.13 & 48.01 & 3.86 \\
      & 100 hyp & 26.40 & 7.31 & \textbf{42.31} & \textbf{66.32} & \textbf{48.32} & 3.91 \\
      \\
      \multirow{1}{*}{VHRED$_{\textit{lgm}100}$} & 100 hyp & 26.25 & \textbf{7.39} & 42.28 & 66.19 & 48.16 & \textbf{3.92} \\
      \midrule
      \multicolumn{8}{c}{\textit{Prior works} (\S\ref{ssec:prior_models}) } \\
      \midrule
      MHRED & 100 hyp & 24.27 & 6.59 & 39.65 & 61.64 & 44.79 & 3.80 \\
      HRED$_{\textit{CVaR}}$ & 100 hyp & 20.92 & 7.32 & 40.49 & 62.43 & 45.53 & 3.88 \\
      VHRED$_{\textit{MBoW}}$ & 100 hyp & 51.74 & 5.68 & 38.71 & 62.81 & 45.07 & 3.41 \\
      \bottomrule
  \end{tabu}
  \caption{Experimental results of model enchancement for multi-referenced training.}
  \label{tab:model_result}
\end{table*}

\subsection{Larger-Sized Model}
\label{ssec:larger_model}
The simplest way to increase a model's capacity is to use more hidden units and layers. Since the baseline HRED has 1 hidden layer with 500 hidden units, we experimented with larger HREDs, which are 1) HRED$_{\textit{l}}$ with 2 layers and 1000 hidden units per layer and 2) HRED$_{\textit{xl}}$ with 2 layers and 2000 hidden units per layer. As shown in Table~\ref{tab:model_result}, HRED$_{\textit{l}}$ slightly outperforms the original HRED but a larger HRED$_{\textit{l}}$ yields worse results in some metrics. It suggests that increasing model size is not a consistent way to improve performance.

\subsection{Variational Model}
\label{ssec:variational_model}
VHRED and VHRED$_{\textit{gmm}}$ have the potential to learn one-to-many relations better since they can generate different output sequences by sampling different values from its encoding distributions. However, their performance is not even comparable with the baseline HRED according to Table~\ref{tab:model_result}. We also found the performance of VHRED and VHRED$_{\textit{gmm5}}$ with larger latent variable size and more components to be worse, which is partially due to the fact that their KLD losses are positively correlated with the latent variable size and thus are unbalanced with their reconstruction losses. These results suggest that existing variational baselines are not expressive enough and difficult to optimize.

\subsection{VHRED with Linear Gaussian Model (LGM) Prior}
\label{ssec:lgm}

To allow for stronger expressiveness, we propose a linear Gaussian model (LGM) prior. Instead of relying on a single Gaussian latent variable, we exploit $K$ Gaussian latent variables $\mathbf{z}_{1}$ to $\mathbf{z}_{K}$ and use their linear combination to encode a dialogue:
\begin{align*}
  \boldsymbol{\mu}_k, \boldsymbol{\sigma}_k, \pi_k &= \tn{MLP}_{\theta, k}(\mathbf{c}) \\
  \mathbf{z}_k &\sim \tn{Gaussian}(\boldsymbol{\mu}_k, \boldsymbol{\sigma}_k^2\boldsymbol{I}) \\
  \mathbf{z} &= \begin{matrix} \sum_{k=1}^{K} \pi_k \mathbf{z}_k \end{matrix},
\end{align*}
and we refer to the VHRED with $K$-variable LGM prior as VHRED$_{\textit{lgm}K}$.

This simple modification significantly improves VHRED's performance according to results in Table~\ref{tab:model_result}. We experimented with $K$ in $\{5, 20, 100\}$ and found the performance improvement to be consistent with more hypotheses and larger $K$.

Regarding how the interaction between a model's expressiveness (i.e. $K$) and the amount of hypotheses affects model performance, we notice that:
\begin{itemize}
  \item When $K$ is small ($K$ = 5), we can hardly obtain performance gain by training with more hypotheses (from 20 to 100).
  \item When we increase $K$ to 20, further performance gain is achievable. It suggests that the performance bottleneck can be widened to allow for learning from more hypotheses.
  \item When we increase $K$ to 100, the performance gap between VHRED$_{\textit{lgm}20}$ and VHRED$_{\textit{lgm}100}$ is very small. It suggests that we may need more hypotheses to exploit the expressiveness of VHRED$_{\textit{lgm}100}$.
\end{itemize}

\subsection{Comparing with Prior Works}
\label{ssec:prior_models}
Three models from prior works are also used for comparison in Table~\ref{tab:model_result}, including the mechanism-aware model (MHRED, \citealp{zhou2017mechanism}), the conditional value-at-risk model designed for learning different dialogue scenarios (HRED$_{\textit{CVaR}}$, \citealp{zhang2018tailored}), and the two-step variational model (VHRED$_{\textit{MBoW}}$, \citealp{qiu2019training}). Their details have been discussed in Section \ref{ssec:expressive_models}. As shown in Table~\ref{tab:model_result}, these models are not competitive in the multi-referenced setting, and two of them cannot even beat the baseline HRED.






\section{Human Evaluation}
\label{sec:human_eval}

\begin{table}[!t]
  \centering
  \begin{tabu} to 1.0\textwidth {lccc}
      \toprule
      \multirow{2}{*}{\textbf{Model}} & \multicolumn{3}{c}{\textbf{Human Scores (in \%)}} \\
      & \textbf{Sensible} & \textbf{Specific} & \textbf{SSA} \\
      \midrule
      \multicolumn{4}{c}{\textit{Trained on 1-GT data}} \\
      \midrule
      HRED & 59.50 & 60.00 & 59.75 \\
      VHRED$_{\textit{gmm5}}$ & 38.50 & 56.00 & 47.25 \\
      VHRED$_{\textit{lgm20}}$ & 52.50 & 63.50 & 58.00 \\
      \midrule
      \multicolumn{4}{c}{\textit{Trained on 100-hypotheses data}} \\
      \midrule
      HRED & 68.50 & 67.00 & 67.75 \\
      VHRED$_{\textit{gmm5}}$ & 44.50 & 66.50 & 55.50 \\
      VHRED$_{\textit{lgm20}}$ & \textbf{72.50} & \textbf{74.00} & \textbf{73.25} \\
      \bottomrule
  \end{tabu}
  \caption{Results of human evaluation on 3 models trained on 2 types of data.}
  \label{tab:human_eval}
\end{table}

Besides automated evaluation, we also conduct human evaluation to provide a more accurate assessment of model performance. We sample 100 dialogues randomly from the test data and generate responses using 3 models (HRED, VHRED$_{\textit{gmm5}}$, VHRED$_{\textit{lgm5}}$) trained on 2 types of data (the 1-GT data and the 100-hypotheses data). We ask 4 Amazon MTurk human workers to annotate the sensibleness and the specificity of the 600 $(context, response)$ pairs. The collected data reach good inter-rater agreement (Krippendorff's $\alpha$ $>$ 0.6). Then we calculate the average of the two metrics (SSA,~\citealp{adiwardana2020towards}) as introduced in Section~\ref{ssec:metrics}. The results of the human evaluation are given in Table~\ref{tab:human_eval}. First, all three models obtain significant improvements on all three metrics by training on the multi-referenced data, which confirms the effectiveness of the proposed data enhancement method. Then, VHRED$_{\textit{lgm20}}$ is better than its GMM counterpart and the HRED. And a larger performance gain is obtained for VHRED$_{\textit{lgm20}}$ than other models when we train it on the multi-referenced data. The result suggests that an expressive prior is indeed necessary and useful for latent dialogue models, especially in the multi-referenced setting.

\begin{table*}[!ht]
  \centering
  \begin{tabu} to 1.0\textwidth {crcccccc}
      \toprule
      \multirow{2}{*}{\textbf{Use GT}} & \multirow{2}{*}{\textbf{\# hyp.}} & \multirow{2}{*}{\textbf{ppl}} & \multirow{2}{*}{\textbf{BLEU-2}} & \multicolumn{3}{c}{\textbf{Embedding Similarity}} & \multirow{2}{*}{\textbf{Reval}} \\
      & & & & \textbf{Ext} & \textbf{Avg} & \textbf{Grd} & \\
      \midrule
      \ding{55} & 1 & 46.45 & 6.65 & 41.10 & 64.95 & 46.77 & 3.64 \\
      \ding{51} & 1 & 30.12 & 6.70 & 41.48 & 65.01 & 46.91 & 3.71 \\
      \ding{55} & 5 & 29.18 & 6.99 & 41.80 & 65.72 & 47.68 & 3.82 \\
      \ding{51} & 5 & 27.31 & 7.26 & 42.21 & 66.33 & 48.32 & 3.83 \\
      \ding{55} & 20 & 26.93 & 7.07 & 42.29 & 66.13 & 48.01 & 3.86 \\
      \ding{51} & 20 & 26.46 & 7.25 & 42.00 & 65.81 & 47.71 & 3.88 \\
      \ding{55} & 100 & 26.40 & 7.31 & 42.31 & 66.32 & 48.32 & 3.91 \\
      \ding{51} & 100 & 26.49 & 7.23 & 42.28 & 65.83 & 47.60 & 3.88 \\
      \bottomrule
  \end{tabu}
  \caption{Experimental results of combining ground truth and hypotheses. (\S\ref{ssec:mix_data})}
  \label{tab:mix_data}
\end{table*}

\section{Analysis}

\subsection{Combining Ground Truth and Hypotheses}
\label{ssec:mix_data}
One issue that readers may be concerned about is whether it is better to combine ground truth with hypotheses than to use them separately. We take the VHRED$_{\textit{lgm}20}$ as an example and conduct experiments using mixed training data. As shown in Table~\ref{tab:mix_data}, we can get performance gain by training with mixed data. The improvement is larger when the original data is smaller (1 hypothesis) because it doubles the training data. When using 100 hypotheses, we can almost fully rely on the generated data and discard ground truth.

\subsection{What do variables in LGM learn?}
\label{ssec:lgm_learn}

\begin{table}[!t]
  \centering
  \begin{tabu} to 1.0\textwidth {ccccc}
      \toprule
      \multirow{2}{*}{\textbf{k}} & \multirow{2}{*}{$\bar{\pi}_k$} & \multirow{2}{*}{\textbf{ppl}} & \multirow{2}{*}{\textbf{BLEU-2}} & \multirow{2}{*}{\textbf{Reval}} \\
      & & & & \\
      \midrule
      \multicolumn{5}{c}{\textit{Bad prob. / bad PPL / bad Reval.}} \\
      \midrule
      4 & 0.12\% & 4865.8 & 1.77 & 1.51 \\
      \midrule
      \multicolumn{5}{c}{\textit{Bad prob. / good PPL / bad Reval.}} \\
      \midrule
      0 & 0.38\% & 112.10 & 5.42 & 2.73 \\
      \midrule
      \multicolumn{5}{c}{\textit{Medium prob. / bad PPL / good Reval.}} \\
      \midrule
      8 & 8.22\% & 2740.2 & 6.22 & 3.74 \\
      \midrule
      \multicolumn{5}{c}{\textit{Good prob. / good PPL / good Reval.}} \\
      \midrule
      1 & 39.24\% & 72.34 & 5.52 & 3.59 \\
      \bottomrule
  \end{tabu}
  \caption{Experimental results of VHRED$_{\textit{lgm}20}$ decoding with the $k$-th latent variable. (\S\ref{ssec:lgm_learn})}
  \label{tab:result_lgm20_k}
\end{table}


\begin{table}[!t]
  \centering
  \tabulinesep=0.7mm
  \begin{tabu} to 1.0\columnwidth {cX[l]}
      \toprule
      \multicolumn{2}{c}{\textbf{Dialogue \#422}} \\
      \midrule
      \textbf{Floor} & \multicolumn{1}{c}{\textbf{Context Utterance}} \\
      \midrule
      A & \textit{i'm so hungry. shall we go eat now, rick?} \\
      B & \textit{sure. where do you want to go? are you in the mood for anything in particular?} \\
      A & \textit{how about some dumplings? i just can't get enough of them.} \\
      B & \textit{[to be predicted]} \\
      \midrule
      \textbf{k} & \multicolumn{1}{c}{\textbf{Response Utterance}} \\
      \midrule
      4 & \textit{tables tables tables there any any any any pale, medium rare.} \\
      0 & \textit{ok. i don't think we have any soup at the moment.} \\
      8 & \textit{i've heard that some dumplings are really good. but i don't know what to eat.} \\
      1 & \textit{ok. i'll go to the restaurant.} \\
      \bottomrule
  \end{tabu}
  \caption{Samples of VHRED$_{\textit{lgm}20}$ decoding with the $k$-th latent variable. (\S\ref{ssec:lgm_learn})}
  \label{tab:sample_lgm20_k}
\end{table}

We combine latent variables linearly in the LGM prior. To investigate how each variable contributes, we train a standard VHRED$_{\textit{lgm}20}$ on the 100-hypotheses data, but evaluate it by using only 1 variable to generate responses. Besides the metrics introduced above, we calculate the average selection probability $\pi_k$ on the test data (as denoted by $\bar{\pi}_k$). Out of the results, we find four obvious patterns regarding their selection probability (avg prob.), perplexity (PPL), and RoBERTa-eval scores (Reval.). The results of these patterns are shown in Table~\ref{tab:result_lgm20_k}.

In general, selection probability correlates positively with RoBERTa-eval score, while perplexity is less relevant to the other two metrics. For variables that have high probabilities and RoBERTa-eval scores (e.g. the 8th and the 1st), there is a performance discrepancy on other metrics, and thus we believe LGM can capture different aspects of responses. For instance, we notice that the 1st variable tends to generate generic and safe responses, while the 8th variable is likely to produce sentences with more diverse word types. A dialogue example is given in Table~\ref{tab:sample_lgm20_k}.\footnote{More examples and results can be found in the Appendix.} More exact interpretation of the variables remains challenging, and we leave this to future works.

\section{Conclusion}
\label{sec:conclusion}
In this work, we analyzed the training objective of dialogue response generation models from the view of distribution distance as measured by Kullback\textendash Leibler divergence. The analysis showed that single-referenced dialogue data cannot characterize the one-to-many feature of open-domain dialogues and that multi-referenced training is necessary. Towards multi-referenced training, we first proposed to enhance the training data by replacing every single reference with multiple hypotheses generated by a finetuned GPT2, which provided us with a better approximation of the real data distribution. Secondly, we proposed to equip variational dialogue models with an expressive prior, named linear Gaussian model (LGM), to capture the one-to-many relations. The automated and human evaluation confirmed the effectiveness of the proposed methods.

\bibliography{collection}
\bibliographystyle{acl_natbib}

\appendix

\section{Data Preprocessing}
\label{sec:appendix_corpus}
For deduplication, we iterate over the sessions and remove those in which more than 50\% utterances have appeared in a seen session. We also regularize the text by converting all the letters into lower case and adjusting punctuation positions (e.g. ``\textit{I ' m good.}'' $\rightarrow$ ``\textit{i 'm good.}''). We further tokenize all the utterances using \texttt{spaCy} English tokenizer.\footnote{\url{https://spacy.io/}} We show the number of sessions and $(context, response)$ pairs in the processed data in Table~\ref{tab:appendix_corpus}.

\section{Human Evaluation}
\label{sec:appendix_human_eval}
We recieved 2400 annotations in total (4 annotators for each of the 600 $(context, response)$ pairs). We first remove annotation outliers following \citet{leys2013detecting}. After removing 208 annotations for sensibleness and 253 for specificity, the remaining annotations have reasonable inter-rater agreement meansured by Krippendorff's $\alpha$~\citep{krippendorff2018content} as shown in Table~\ref{tab:appendix_agreement}.

\section{Experimental Settings}

\subsection{Model Implementation}
\label{ssec:appendix_model}
For HRED and VHRED models, we implement encoders and decoders with gated recurrent unit (GRU) networks. Sentence-level encoders are bidirectional, while dialogue-level encoders and decoders are unidirectional. All the GRU networks have 1 layer and 500 hidden units. We use 30-dimensional floor embeddings to encode the switch of floor. For VHREDs, latent variables have 200 dimensions. Prior and posterior networks are implemented by feedforward networks with hyperbolic tangent activation function. While priors have different forms (unimodal Gaussian, Gaussian mixture model, and linear Gaussian model), we use unimodal Gaussian for all the posteriors. We use attentional mechanism for all decoders. In Table~\ref{tab:appendix_model_size}, we show the number of model parameters and training time per epoch on the 1 ground-truth data using a single \texttt{NVIDIA TITAN RTX} card. When training on $K$-hypotheses data, the training time per epoch is roughly $K$ times of the reported number.

\subsection{Training Details}
\label{ssec:appendix_training}
We optimize all the models with the Adam method~\citep{kingma2014adam}. The initial learning rate is 0.001 and gradients are clipped within [-1.0, 1.0]. We decay the learning rate with decay rate 0.75 and patience 3. The training process is early stopped when the learning rate is less than 1$\times$10$^{-7}$. The numbers of training epochs and steps are shown in Table~\ref{tab:appendix_epoch}. Batch size is 30 during training. We use up to 5 history utterances as context, and all utterances are truncated to have 40 tokens to most. We set dropout probability as 0.2 and shuffle training data every epoch for better generalization. VHREDs are optimized by maximizing their variational lower bound~\citep{sohn2015learning}. We apply linear KL annealing in the first 40,000 training steps. 

For finetuning the GPT2 model, we use a smaller batch size of 10 to fit the model into memory. As with other hyperparameters such as learning rate and weight regularization factor, we follow the settings used by~\citet{wolf2019transfertransfo}. And the GPT2 is finetuned on the 1-GT data for only 2 epochs.

\begin{table}[!t]
  \centering
  \begin{tabu} to 1.0\textwidth {lc}
      \toprule
      \textbf{Item} & \textbf{Krippendorff's $\alpha$} \\
      \midrule
      Sensibleness & 0.76 \\
      Specificity & 0.60 \\
      \bottomrule
  \end{tabu}
  \caption{Inter-rater agreement of human annotations. (\S~\ref{sec:appendix_human_eval})}
  \label{tab:appendix_agreement}
\end{table}

\begin{table}[!t]
  \centering
  \begin{tabu} to 1.0\textwidth {lrr}
      \toprule
      \textbf{Model} & \textbf{\# Parameters} & \textbf{Trn. Time}\\
      \midrule
      HRED & 8.04 M & $\sim$150 s \\
      VHRED & 11.02 M & $\sim$160 s \\
      VHRED$_{\textit{gmm}}5$ & 11.36 M & $\sim$160 s \\
      VHRED$_{\textit{lgm}}5$ & 11.36 M & $\sim$160 s \\
      VHRED$_{\textit{lgm}}20$ & 12.52 M & $\sim$160 s \\
      VHRED$_{\textit{lgm}}100$ & 18.67 M & $\sim$160 s \\
      GPT2$_{\textit{md}}$ & 338.39 M & $\sim$3000 s \\
      \bottomrule
  \end{tabu}
  \caption{Number of parameters and training time per epoch for each model. (\S~\ref{ssec:appendix_model})}
  \label{tab:appendix_model_size}
\end{table}

\begin{table}[!t]
  \centering
  \begin{tabu} to 1.0\textwidth {lccc}
      \toprule
      \multirow{2}{*}{\textbf{Item}} & \multicolumn{3}{c}{\textbf{Statistics}} \\
      & \textit{Train} & \textit{Validation} & \textit{Test} \\
      \midrule
      sessions & 9237 & 1157 & 1159 \\
      $(ctx, resp)$ pairs & 59305 & 9906 & 9716 \\
      \bottomrule
  \end{tabu}
  \caption{Corpus statistics. (\S~\ref{sec:appendix_corpus})}
  \label{tab:appendix_corpus}
\end{table}

\begin{table}[!t]
  \centering
  \begin{tabu} to 1.0\textwidth {rrr}
      \toprule
      \textbf{Training Data} & \textbf{Max Epochs} & \textbf{Max Steps} \\
      \midrule 
      1 GT & 100 & 5.93M \\
      1 hyp. & 100 & 5.93M \\
      1 GT + 1 hyp. & 50 & 5.93M \\
      5 hyp. & 20 & 5.93M \\
      1 GT + 5 hyp. & 20 & 7.12M \\
      20 hyp. & 10 & 11.86M \\
      1 GT + 20 hyp. & 10 & 12.45M \\
      100 hyp. & 2 & 11.86M \\
      1 GT + 100 hyp. & 2 & 11.98M \\
      \bottomrule
  \end{tabu}
  \caption{Maximum training epochs and steps in different data settings. (\S~\ref{ssec:appendix_training})}
  \label{tab:appendix_epoch}
\end{table}

\section{Extra Results}
\label{sec:appendix_extra_results}

In Table~\ref{tab:appendix_result_lgm20_k}, we presented the full results of experiments in Section~\ref{ssec:lgm_learn}.

\begin{table*}[!t]
  \centering
  \begin{tabu} to 1.0\textwidth {cccccccccc}
      \toprule
      \multirow{2}{*}{\textbf{k}} & \multirow{2}{*}{\textbf{avg prob.}} & \multirow{2}{*}{\textbf{PPL}} & \multirow{2}{*}{\textbf{B-2}} & \multicolumn{3}{c}{\textbf{Emb. Sim.}} & \multirow{2}{*}{\textbf{Dist-1}} & \multirow{2}{*}{\textbf{Dist-2}} & \multirow{2}{*}{\textbf{Reval}} \\
      & & & & \textbf{E} & \textbf{A} & \textbf{G} & & & \\
      \midrule
      \multicolumn{10}{c}{\textit{VHRED$_{\textit{lgm}20}$ using all variables}} \\
      \midrule
      mix & 100.00 & 26.40 & 7.31 & 42.31 & 66.32 & 48.32 & 1677 & 7641 & 3.91 \\
      \midrule
      \multicolumn{10}{c}{\textit{VHRED$_{\textit{lgm}20}$ using the $k$-th variable}} \\
      \midrule
      0 & 0.38 & 112.10 & 5.42 & 37.05 & 60.24 & 45.72 & 1371 & 6791 & 2.73 \\
      1 & 39.24 & 72.34 & 5.52 & 40.75 & 62.76 & 43.88 & 1096 & 4291 & 3.59 \\
      2 & 4.78 & 62.70 & 5.93 & 38.71 & 65.16 & 50.01 & 2297 & 12635 & 3.74 \\
      3 & 2.63 & 55.96 & 6.93 & 40.37 & 66.71 & 51.05 & 2191 & 12271 & 3.85 \\
      4 & 0.12 & 4865.8 & 1.77 & 30.70 & 50.97 & 38.04 & 1827 & 20293 & 1.51 \\
      5 & 0.70 & 4064.8 & 5.68 & 39.10 & 63.99 & 45.66 & 1557 & 8065 & 3.15 \\
      6 & 5.12 & 3491.0 & 6.29 & 40.03 & 65.88 & 49.93 & 2228 & 12977 & 3.72 \\
      7 & 5.38 & 3078.1 & 5.86 & 40.50 & 63.48 & 45.14 & 2006 & 10773 & 3.39 \\
      8 & 8.22 & 2740.2 & 6.22 & 39.11 & 64.61 & 48.24 & 1999 & 9891 & 3.74 \\
      9 & 1.49 & 2469.5 & 7.08 & 41.77 & 65.91 & 47.83 & 1706 & 8056 & 3.76 \\
      10 & 0.24 & 2788.9 & 2.29 & 28.75 & 45.30 & 32.57 & 1175 & 9400 & 1.83 \\
      11 & 0.21 & 2855.4 & 2.67 & 31.74 & 50.69 & 35.67 & 925 & 6663 & 1.75 \\
      12 & 0.17 & 13705 & 1.19 & 28.90 & 46.67 & 34.29 & 2563 & 27438 & 1.59 \\
      13 & 2.25 & 12729 & 6.27 & 39.51 & 66.81 & 52.20 & 2310 & 13391 & 3.81 \\
      14 & 0.23 & 12847 & 1.92 & 30.24 & 50.94 & 37.92 & 2668 & 23246 & 1.73 \\
      15 & 13.17 & 12046 & 6.74 & 42.05 & 65.34 & 46.95 & 1554 & 6753 & 3.88 \\
      16 & 0.24 & 12998 & 1.58 & 29.97 & 49.45 & 36.48 & 1770 & 17281 & 1.66 \\
      17 & 0.22 & 12331 & 3.50 & 33.43 & 55.81 & 41.57 & 1269 & 7089 & 2.07 \\
      18 & 0.26 & 11684 & 7.27 & 42.25 & 66.69 & 48.55 & 1649 & 7909 & 3.79 \\
      19 & 14.90 & 11101 & 6.60 & 41.68 & 65.23 & 47.00 & 1533 & 6625 & 3.84 \\
      \bottomrule
  \end{tabu}
  \caption{Experimental results of VHRED$_{\textit{lgm}20}$ decoding with the $k$-th latent variable.}
  \label{tab:appendix_result_lgm20_k}
\end{table*}

\section{Extra Samples}
\label{sec:appendix_sample}

We show three more dialogue samples in Table~\ref{tab:appendix_sample1},~\ref{tab:appendix_sample2}, and~\ref{tab:appendix_sample3}. Notice that we have collapsed some spaces between the letters and the punctuation marks for easy reading.

\begin{table*}[!t]
  \centering
  \tabulinesep=0.7mm
  \begin{tabu} to 1.0\textwidth {cX[l]}
    \toprule
    \multicolumn{2}{c}{\textbf{Dialogue \#100}} \\
    \midrule
    \textbf{Floor} & \multicolumn{1}{c}{\textbf{Context Utterance}} \\
    \midrule
    A & \textit{i have never shopped there, but i think price club is a kind of membership store.}  \\
    B & \textit{what does that mean?} \\
    A & \textit{that means you have to pay a membership fee to shop there. i have heard they have very good prices on electronics.} \\
    B & \textit{electronics?} \\
    A & \textit{yes. like stereos and [UNK].} \\
    B & \textsc{[To be predicted]} \\
    \midrule
  \end{tabu}
  \begin{tabu} to 1.0\textwidth {lrX[l]}
      \textbf{Model} & \textbf{Trn. Data} & \multicolumn{1}{c}{\textbf{Response Utterance}} \\
      \midrule
      Ground truth & - & \textit{how much is the membership fee?} \\
      HRED & 1 GT & \textit{do you have any other advantages?} \\
      VHRED$_{\textit{gmm}5}$ & 1 GT & \textit{how do you want to buy?} \\
      VHRED$_{\textit{lgm}20}$ & 1 GT & \textit{so what? i don't have a current account.} \\
      HRED & 100 hyp. & \textit{i see. i'll buy a new one.} \\
      VHRED$_{\textit{gmm}5}$ & 100 hyp. & \textit{that's a good idea. how about the one?} \\
      VHRED$_{\textit{lgm}20}$ & 100 hyp. & \textit{i see. is there a warranty?} \\
      \quad - the 4th variable & 100 hyp. & \textit{sure. which service do you use?} \\
      \quad - the 0th variable & 100 hyp. & \textit{how much money can you get it?} \\
      \quad - the 8th variable & 100 hyp. & \textit{i have a membership card. it's very expensive.} \\
      \quad - the 1st variable & 100 hyp. & \textit{i see. it's a good deal.} \\
      \bottomrule
  \end{tabu}
  \caption{More samples on the test data. (\S\ref{sec:appendix_sample})}
  \label{tab:appendix_sample1}
\end{table*}

\clearpage

\begin{table*}[!t]
  \centering
  \tabulinesep=0.7mm
  \begin{tabu} to 1.0\textwidth {cX[l]}
    \toprule
    \multicolumn{2}{c}{\textbf{Dialogue \#500}} \\
    \midrule
    \textbf{Floor} & \multicolumn{1}{c}{\textbf{Context Utterance}} \\
    \midrule
    A & \textit{i am not sleepy.}  \\
    B & \textit{it's getting late, and you have to wake up early tomorrow.} \\
    A & \textit{there is no way that i can fall asleep right now.} \\
    B & \textit{try listening to some soft music.} \\
    A & \textit{it won't work. i'm nowhere close to being tired.} \\
    B & \textsc{[To be predicted]} \\
    \midrule
  \end{tabu}
  \begin{tabu} to 1.0\textwidth {lrX[l]}
      \textbf{Model} & \textbf{Trn. Data} & \multicolumn{1}{c}{\textbf{Response Utterance}} \\
      \midrule
      Ground truth & - & \textit{i really don't care, just go to sleep.} \\
      HRED & 1 GT & \textit{you're too tired.} \\
      VHRED$_{\textit{gmm}5}$ & 1 GT & \textit{what's up?} \\
      VHRED$_{\textit{lgm}20}$ & 1 GT & \textit{you shouldn't have to sleep.} \\
      HRED & 100 hyp. & \textit{don't worry. i'll get you up.} \\
      VHRED$_{\textit{gmm}5}$ & 100 hyp. & \textit{i don't know. i just want to relax.} \\
      VHRED$_{\textit{lgm}20}$ & 100 hyp. & \textit{you should be tired. you can get a good sleep.} \\
      \quad - the 4th variable & 100 hyp. & \textit{do do let you gift you gift you gift you live you live here i sing for here friendship akimbo?} \\
      \quad - the 0th variable & 100 hyp. & \textit{don't be at the evening.} \\
      \quad - the 8th variable & 100 hyp. & \textit{you are always sleepy.} \\
      \quad - the 1st variable & 100 hyp. & \textit{come on. you can get a good sleep.} \\
      \bottomrule
  \end{tabu}
  \caption{More samples on the test data. (\S\ref{sec:appendix_sample})}
  \label{tab:appendix_sample2}
\end{table*}

\clearpage

\begin{table*}[!t]
  \centering
  \tabulinesep=0.7mm
  \begin{tabu} to 1.0\textwidth {cX[l]}
    \toprule
    \multicolumn{2}{c}{\textbf{Dialogue \#2500}} \\
    \midrule
    \textbf{Floor} & \multicolumn{1}{c}{\textbf{Context Utterance}} \\
    \midrule
    A & \textit{the style of the exterior is just what i've been looking for!} \\
    B & \textit{the exterior is only half of the story. the interior is incredible.} \\
    A & \textit{this home is gorgeous on the inside!} \\
    B & \textit{everything that you see on the walls, floors, and windows is brand new.} \\
    A & \textit{the wall colors totally suit the style of this house.} \\
    B & \textsc{[To be predicted]} \\
    \midrule
  \end{tabu}
  \begin{tabu} to 1.0\textwidth {lrX[l]}
      \textbf{Model} & \textbf{Trn. Data} & \multicolumn{1}{c}{\textbf{Response Utterance}} \\
      \midrule
      Ground truth & - & \textit{straight ahead is the [UNK] kitchen.} \\
      HRED & 1 GT & \textit{i think it's a good idea.} \\
      VHRED$_{\textit{gmm}5}$ & 1 GT & \textit{it's [UNK]! the carpet is behind the curtain.} \\
      VHRED$_{\textit{lgm}20}$ & 1 GT & \textit{you can have a look at the kitchen.} \\
      HRED & 100 hyp. & \textit{i'm glad you like it.} \\
      VHRED$_{\textit{gmm}5}$ & 100 hyp. & \textit{this one is the latest design.} \\
      VHRED$_{\textit{lgm}20}$ & 100 hyp. & \textit{the color is perfect. i can see it.} \\
      \quad - the 4th variable & 100 hyp. & \textit{any any all in all all all any part part of mind?} \\
      \quad - the 0th variable & 100 hyp. & \textit{the kitchen is very nice. i think the color is perfect.} \\
      \quad - the 8th variable & 100 hyp. & \textit{look, the walls are beautiful.} \\
      \quad - the 1st variable & 100 hyp. & \textit{i know. it's a good idea.} \\
      \bottomrule
  \end{tabu}
  \caption{More samples on the test data. (\S\ref{sec:appendix_sample})}
  \label{tab:appendix_sample3}
\end{table*}

\end{document}